\documentclass{article}

\usepackage{microtype}
\usepackage{graphicx}
\usepackage{subcaption}
\usepackage{booktabs}

\usepackage{hyperref}

\usepackage[accepted]{icml2026}

\usepackage{amsmath}
\usepackage{amssymb}
\usepackage{mathtools}
\usepackage{amsthm}

\usepackage[capitalize,noabbrev]{cleveref}

\theoremstyle{plain}

\theoremstyle{definition}

\theoremstyle{remark}

\usepackage{booktabs}
\usepackage{tabularx}
\usepackage{enumitem}
\usepackage{array}
\usepackage{titlesec}
\usepackage{xcolor}
\usepackage{cleveref}
\usepackage{listings}

\usepackage[most]{tcolorbox}
\usepackage{fvextra}

\definecolor{promptbg}{HTML}{F8F8F8}
\definecolor{promptframe}{HTML}{CCCCCC}

\lstdefinestyle{promptstyle}{
  basicstyle        = \ttfamily\fontsize{7pt}{9.5pt}\selectfont,
  breaklines        = true,
  breakatwhitespace = false,
  breakindent       = 0pt,
  frame             = single,
  framesep          = 5pt,
  rulecolor         = \color{promptframe},
  backgroundcolor   = \color{promptbg},
  xleftmargin       = 8pt,
  xrightmargin      = 8pt,
  keepspaces        = true,
  columns           = flexible,
  showstringspaces  = false,
  tabsize           = 2,
  captionpos        = b,
  aboveskip         = 8pt,
  belowskip         = 4pt
}

\setlist{
  itemsep=2pt,      
  parsep=0pt,       
  topsep=2pt,       
  partopsep=0pt
}

\titlespacing*{\paragraph}
  {0pt}    
  {2pt}    
  {9pt}    

\newcommand{\Succeed}{\boldsymbol{\mathrm{Succeed}}}
\newcommand{\Fail}{\boldsymbol{\mathrm{Fail}}}
\newcommand{\Plus}{\boldsymbol{\mathrm{+}}}
\newcommand{\Minus}{\boldsymbol{\mathrm{-}}}

\newcommand{\SucceedPlus}{\boldsymbol{\mathrm{Succeed}^{+}}}
\newcommand{\FailPlus}{\boldsymbol{\mathrm{Fail}^{+}}}
\newcommand{\FailMinus}{\boldsymbol{\mathrm{Fail}^{-}}}

\usepackage{xcolor}

\definecolor{boxbg}{HTML}{FAFAFA}
\definecolor{boxframe}{HTML}{D0D7DE}
\definecolor{boxtitle}{HTML}{F3F4F6}

\tcbset{
  enhanced,
  breakable,
  boxrule=0.5pt,
  arc=1.2pt,
  colback=boxbg,
  colframe=boxframe,
  left=6pt,
  right=6pt,
  top=5pt,
  bottom=5pt
}

\newtcolorbox{artifactbox}[2][]{
  title=\textbf{#2},
  colbacktitle=boxtitle,
  coltitle=black,
  fonttitle=\small,
  breakable,
  #1
}

\newtcolorbox{promptbox}[2][]{%
  enhanced,
  breakable,
  colback=gray!4!white,
  colframe=gray!55!black,
  boxrule=0.5pt,
  arc=1.5pt,
  fonttitle=\bfseries\small,
  coltitle=black,
  title=#2,
  attach boxed title to top left={yshift=-2mm, xshift=3mm},
  boxed title style={colback=white, colframe=gray!55!black, sharp corners, boxrule=0.4pt},
  left=5pt, right=5pt, top=6pt, bottom=5pt,
  before upper={\small\setlength{\parskip}{3pt}},
  #1
}

\icmltitlerunning{Global Merger-Arbitrage Forecasting with Language Models}

\begin{document}

\twocolumn[
  \icmltitle{Global Merger-Arbitrage Forecasting with Language Models}

  \icmlsetsymbol{equal}{*}

  \begin{icmlauthorlist} 
    \icmlauthor{Hinal Jajal}{comp}
    \icmlauthor{Michal Mucha}{comp}
    \icmlauthor{Charles Sweat}{comp}
    \icmlauthor{Chris Pulman}{comp}
    \icmlauthor{Charlie Flanagan}{comp}
    \icmlauthor{Peter Anderson}{comp}
  \end{icmlauthorlist}

  \icmlaffiliation{comp}{Balyasny Asset Management}

  \icmlcorrespondingauthor{Hinal Jajal}{bamappliedai@bamfunds.com}

  \icmlkeywords{Machine Learning, ICML}

  \vskip 0.3in
]

\printAffiliationsAndNotice{}  

\begin{abstract}

We present a language-model forecasting system for merger arbitrage, a specialized high-stakes financial setting in which the task is to predict the outcome of announced M\&A deals. Unlike prior work on judgmental forecasting with LLMs, which has focused on broad mixed-topic benchmarks and short context such as news snippets, we study a setting that requires long-context reasoning over hundreds of pages of technical documents. Our system combines expert-guided context engineering with finetuning on hindsight-guided reasoning traces derived from historical deals. Given an announced deal, it outputs a probability distribution over three mutually exclusive outcomes: closing at announced terms, a higher bid, or deal termination. 
On an out-of-sample set of more than 400 large deals spanning 42 countries, our finetuned system achieves the best performance of any method we evaluate, reducing class-balanced Brier score to 0.151. This is 24\% below calibrated market-implied probabilities, 19\% below XGBoost, and 25-42\% below frontier language models.
These results, together with ablation studies, show that LLM-based forecasting can succeed in specialized, long-context financial workflows, with hindsight-based supervision and expert-designed context playing a critical role.

\end{abstract}

\section{Introduction}

Judgmental forecasting---probabilistic prediction of future events based on documents and contextual reasoning, rather than explicit statistical models---is increasingly being studied in the context of large language models. Recent work shows that LLMs can produce calibrated probability forecasts approaching the accuracy of some human forecasters \cite{halawi2024approaching,schoen2024wisdom,alur2025aia}. However, this literature largely focuses on broad, mixed-topic question banks, posed with only shallow context such as news headlines and blurbs \cite{kargerforecastbench}. It remains unclear whether LLM-based forecasting can add value in specialized, high-stakes domains that require analyzing long and highly technical texts.

We focus on a financial domain: forecasting the outcomes of announced mergers and acquisitions (M\&A). Such forecasts are actively monetized in merger-arbitrage strategies, where an investor buys the target firm’s stock after a deal is announced (possibly against a short position in the acquirer) to capture the spread between the current price and the deal consideration. The investor earns this spread if the deal closes and faces losses if the deal terminates. Therefore, accurate probability forecasts of deal closure and alternative resolutions are central to position sizing, risk management, and portfolio construction.

Arriving at these probability forecasts requires processing large volumes of complex text. Analyzing a single deal can require reading a hundred-page merger agreement; assessing competitive overlaps, regulatory regimes and political considerations across jurisdictions to anticipate regulatory responses; comparing against outcomes in similar past transactions; scrutinizing shareholder bases, voting histories, and public statements; reviewing risks in the acquirer’s financing and balance sheet, and more. At the same time, analysts must integrate this information with a continuous stream of regulatory filings, press releases, and expert analyses.

We build an LLM-based forecasting system tailored to this setting. First, retrieval-augmented research agents each gather and analyze a specific dimension of deal context. Second, an ensembled, finetuned frontier model uses the resulting in-depth context (typically 6.8-10k tokens) to generate a probabilistic forecast and a detailed deal report. The research agents are optimized over months of collaboration with veteran merger-arb specialists.
The frontier model is finetuned on gold training targets derived from post-mortem analysis. For each historical deal, we generate hindsight-guided forecasts at multiple points in time to identify the evidence that should have been weighted most heavily given the realized outcome.
We enforce strict temporal integrity, for example by excluding open-web sources prone to date-filter leakage, and use only models with knowledge cutoffs preceding the test period. This process yields quality reasoning traces that train the model to reason about deal risk and generate meaningful reports. 

On a held-out test set of 404 large deals spanning 42 countries, our finetuned system outperforms frontier models, probabilities inferred from market prices, and an XGBoost model trained on a rich set of deal features across a range of weighted Brier scores. Its best class-balanced Brier score is 0.151, compared with 0.199 for Platt-scaled market-implied probabilities, 0.186 for XGBoost, and 0.201--0.259 for frontier models given identical context (lower is better). Finetuning drives much of this gain, reducing calibration error from 0.089 to 0.039. More broadly, LLM-based approaches are less tied to market consensus than XGBoost: the correlation with market-implied probabilities is 0.36 for our finetuned system, compared with 0.75 for XGBoost.

Ablation studies show that each component contributes materially to performance: class-balanced Brier score worsens by 0.032 without hindsight guidance, by 0.021 when the system is limited to Deal Card information, and by 0.026 when training on only half of the 1244-deal training set. On the basis of these results, we deploy the system as a decision-support tool for analysts and portfolio managers.

\section{Related Work}

\citet{halawi2024approaching} are the first to demonstrate that LLMs, when provided with retrieved news context, can approach non-expert human performance in backtests on broad prediction-market question sets—especially early in the forecast, when human forecasters are uncertain and at least five relevant news articles are available. Building on this, \citet{hsieh2024reasoning} augment LLM forecasters with a ReAct \cite{yao2022react} loop that incorporates web search and Python tools. 
In a complementary direction, \citet{schoen2025augmented} show that LLM assistants improve human forecasting accuracy.

More recently, AIA Forecaster \cite{alur2025aia} focuses on agentic search, forecast calibration, and ensembling with a supervisor agent, matching expert human performance on ForecastBench. They also introduce MarketLiquid, constructed from liquid prediction-market contracts, and show that while their system alone underperforms the market-implied probabilities, an ensemble that combines AIA Forecaster with market prices yields forecasts that are more accurate than market prices alone.

Given the many potential sources of information leakage in backtesting LLM forecasters \cite{paleka2025pitfalls}, recent work has focused on standardizing and improving evaluation protocols. \citet{sudhir2024consistency} propose a suite of logical consistency checks. Bench to the Future \cite{wildman2025bench} mitigates contamination by providing a per-question snapshotted web corpus. To address the erosion of static benchmarks as LLM knowledge cutoffs advance, ForecastBench \cite{kargerforecastbench} continuously aggregates unresolved questions from prediction platforms and prediction markets, enabling evaluation of humans and LLMs. \citet{daillms} introduce a continuous evaluation that uses daily news as an oracle, automatically generating forecasting questions and showing that accuracy degrades smoothly as evaluation dates drift beyond training cutoffs, even with retrieval-augmented generation.

Complementing these backtests, small-scale live evaluations have been reported. \citet{schoen2024wisdom} evaluate 12 LLMs on 31 prediction market questions within 48 hours of question opening, finding no statistical difference in performance between the human crowd and the LLM crowd. \citet{alur2025aia} also report a small-scale live evaluation that is competitive with prediction markets.

Across all evaluation protocols—static, dynamic, and live—prior work has converged on mixed sets of heterogeneous questions. We argue that investigating \emph{only broad, mixed-topic question banks} is a mistake. Combining sports, geopolitics, climate, macroeconomics, politics, and more collapses qualitatively different forecasting problems into a single score, obscuring where models succeed or fail, providing poor guidance for model and system development, and incentivizing shallow, generic retrieval and prompting. This setup also systematically ignores the domain-specific data, tools, and workflows that define specialist forecasting in practice, and therefore risks overstating the readiness of current LLM forecasters for deployment.

In contrast to previous work, we study a single, economically important domain in depth and design our system around the artifacts and workflows of specialist forecasters. Rather than treating retrieval as generic web search over news snippets, we build a retrieval and reasoning pipeline targeted at long, technical financial documents. Methodologically, aside from \citet{halawi2024approaching} and \citet{turtel2025outcome}, there is little work on finetuning to improve judgmental forecasts, and thus limited understanding of how to construct gold targets. We address this gap and demonstrate the value of finetuning hindsight into the model.

\section{Task and Dataset}

\subsection{Task Definition}

We aim to predict the outcome of an announced merger or acquisition. The prediction has three components:

\paragraph{1. Deal Outcome Probabilities.} We distinguish between three mutually exclusive outcomes:
\begin{enumerate}
    \item $\SucceedPlus$: The deal closes as announced.
    \item $\FailPlus$: The announced deal is terminated, but the target shareholders experience a \emph{positive} outcome as a higher bid emerges.
    \item $\FailMinus$: The announced deal is terminated with a \emph{negative} outcome for target shareholders (no higher bid).
\end{enumerate}
Differentiating between positive and negative terminated deals is crucial for commercializing predictions, and for comparing to market-implied probability (refer Sec. \ref{sec:implied-prob}).
\paragraph{2. Days to Completion.} Prediction of the days to deal completion, assuming the deal closes as announced.
\paragraph{3. Deal Report.} A detailed explanation of the reasoning underlying the predicted probability distribution, including citations and key deal-specific risks and mitigants to monitor, such as a termination fee or a pre-existing lawsuit. Identifying key risks in the deal is central to merger-arb, especially for discretionary---rather than systematic---traders. 

\subsection{Dataset Construction}

There is no readily available dataset of historical M\&A deals and their outcomes. We construct a historical sample of 1,648 public-target M\&A deals 
covering the 4 year period from 1 January 2022 to 31 December 2025.
This horizon balances the need for a sufficiently deep history to support finetuning against the difficulty of reconstructing rich deal context from older text sources.

We exclude transactions that do not represent classic, control-oriented merger-arb situations---specifically, minority-stake acquisitions ($\le 50\%$), asset sales, and SPAC transactions---where there is no clean, tradable takeover spread for target shareholders. We also exclude small deals (e.g. $<$US\$1bn deal value) that lack institutional liquidity.

Several key fields in the original dataset are incomplete. We use an ensemble of LLM-based ReAct \cite{yao2022react} agents, with access to the search tools defined in Sec. \ref{sec:input-data}, to enrich each deal with the following metadata:

\begin{itemize}
    \item Announcement date and resolution date.
    \item Deal terms and their revision history (e.g., consideration type, currency, cash per share, exchange ratio).
    \item Company guidance on expected closing timing and subsequent updates.
    \item Presence and sequence of competing bids.
\end{itemize}
\noindent
Incomplete or ambiguous information is reviewed manually. 

\subsubsection{Market-Implied Probability}
\label{sec:implied-prob}

After deal announcement, the target’s stock price incorporates investors’ expectations about deal resolution, allowing us to infer the \emph{market-implied probability} of a favorable outcome for target shareholders. We use this both as a training signal for the model and as an input to baselines for comparison on out-of-sample deals.

Let $S_t$ denote the target’s share price on day $t$ after the announcement. Investors’ valuation is approximated as a two-state mixture between:
\begin{itemize}
    \item an upside value $S^+_t$ corresponding to a positive outcome ($\SucceedPlus$ or $\FailPlus$), and
    \item a downside value $S^-_t$ corresponding to a negative outcome ($\FailMinus$).
\end{itemize}
Under this model,
$$
S_t \approx p^m_t S^+_t + (1 - p^m_t) S^-_t,
$$
where $p^m_t$ is the market-implied probability of a positive outcome. 
This two-state mixture representation of the target’s payoff is standard in derivative and event-driven valuation; see, for example, \citet{hull2018options}.
Solving for $p^m_t$ yields
$$
p^m_t \;=\; \operatorname{clamp}_{[0,1]}\!\left(\frac{S_t - S^-_t}{S^+_t - S^-_t}\right)
$$

For cash deals, $S^+_t$ is constant and equal to the per-share cash consideration, discounted by the risk-free rate over the expected time-to-close inferred from company guidance. When no company guidance is available, we use the median days-to-close for U.S. public deals: 175 days. For stock or mixed deals, $S^+_t$ is the per-share value implied by the announced terms at time $t$ (e.g., exchange ratio times the acquirer’s share price, plus the discounted value of any cash component). The value $S^+_t$ reflects the prevailing deal terms at time $t$, and any amendments (e.g. an increase in the offer price) result in an adjustment to $S^+_t$.

We model the downside $S^-_t$, the expected target price if the deal fails, by allowing standalone value to evolve with market movements scaled by the target’s pre-announcement beta $\beta$:
$$
S^-_t = S^-_0 \exp\bigl(\beta \, r_{m,[0,t]}\bigr),
$$
where $S^-_0$ is the 20-day pre-announcement average price and $r_{m,[0,t]}$ is the cumulative log return of the market index from announcement to day $t$. While this specification captures the effect of prevailing market conditions, realized post-break prices also depend on the reason for termination and the potential for alternative transactions, and therefore vary substantially in practice. Consequently, market-implied probability is informative but should not be treated as definitive; in fact, it benefits from further calibration through Platt scaling (refer Sec. \ref{sec:results}).

\begin{table*}[t]
\caption{Dataset splits, outcome distribution, and total forecast instances across time.}
\label{tab:dataset-splits}
\centering
\small
\begin{tabularx}{\textwidth}{@{}X*{3}{c}*{2}{c}c@{}}
\toprule
\textbf{Split} & \multicolumn{3}{c}{\textbf{Deal Outcomes (\%)}}
& \multicolumn{1}{c}{\textbf{Total Deals}}
& \multicolumn{1}{c}{\textbf{Total Forecasts}}
& \textbf{Time Period} \\
\cmidrule(lr){2-4}
& $\SucceedPlus$ & $\FailPlus$ & $\FailMinus$
& \textbf{\#}
& \textbf{\#}
& \\
\midrule
Train      & 84.4 & 2.2 & 13.4 & 848 & 2032 & Jan 2022 -- Jan 2024 \\
Validation & 82.8 & 3.6 & 13.6 & 396 & 1008 & Feb 2024 -- Jan 2025 \\
Test       & 84.0 & 3.0 & 14.4 & 404 & 1115 & Feb 2025 -- Dec 2025 \\
\midrule
\textbf{Overall} & 84.0 & 2.7 & 13.7 & 1648 & 4155 & Jan 2022 -- Dec 2025 \\
\bottomrule
\end{tabularx}
\end{table*}

\subsubsection{Forecast Dates and Temporal Splits}

We partition deals into training, validation, and test sets based on announcement date, using temporal cutoffs of \emph{31 January 2024} and \emph{31 January 2025}. For each deal, we generate multiple forecast dates between the announcement date and the company-guided expected close date. If no guided close date is available, we instead sample forecast dates at two-month intervals up to the actual close date. All forecast dates are constrained to lie within the temporal bounds of their respective data split.

We use the validation set for hyperparameter selection. For final model evaluation, we retrain on the union of the training and validation sets and report performance on the held-out test set. Table~\ref{tab:dataset-splits} summarizes the data splits, including the number of deals, forecast instances, and outcomes.

\subsection{Information Leakage}
\label{sec:leakage}

\citet{paleka2025pitfalls} highlight several leakage mechanisms that can contaminate LLM forecasting backtests. Our setting mitigates many of these issues.

\paragraph{Model knowledge cutoffs.} We only evaluate LLMs whose knowledge cutoff precedes 31 January 2025 (the start of the test period). The same constraint applies to the embedding and reranking models used for retrieval.
    
\paragraph{System-prompt leakage.} Hidden vendor prompts can leak contemporaneous facts (e.g., the current U.S. president), contaminating forecasting benchmarks. In our domain, such prompts are unlikely to reveal deal-specific outcomes.
    
\paragraph{Piggybacking on human forecasts.} LLMs may copy human forecasts leaked or deliberately admitted into retrieval context, thereby achieving the appearance of near-human performance. In our setting, both expert commentary and documents that mention stock prices or deal spreads can appear in context, which can give the model an implicit read on market-implied probabilities. This is not, by itself, a flaw: our primary baselines rely on market-implied probability directly. In Sec. \ref{sec:results} we show that our system significantly outperforms market implied probability and is less correlated with the market than the XGBoost baseline.
    
\paragraph{Benchmark gaming via correlated questions.} Generic question banks can include many correlated risks that enable gaming \cite{sempere2021alignment}. In contrast, M\&A outcomes are largely idiosyncratic at the deal level.
    
\paragraph{Question selection.} Backtests can leak information through question selection (e.g., conditioning on resolution by a known evaluation date, or biasing towards events that have actually occurred). We reduce this risk by constructing the universe mechanistically from public-target deals.
    
\paragraph{Date-restricted retrieval.} Enforcing temporal cutoffs is notoriously difficult with web search or news APIs: pages are continuously edited, news stories are updated, and search providers inject or scrape real-time widgets that bypass nominal date filters \cite{alur2025aia}. We avoid this class of leakage entirely by explicitly excluding news feeds and open-web content. Instead, we restrict context to curated and timestamped regulatory filings, company releases, and other commercial data sources (refer Sec.~\ref{sec:input-data}).

\paragraph{Hindsight-guided finetuning.} Section~\ref{sec:gold-reasoning} describes our approach to finetuning on hindsight-guided deal reports. The finetuned model is exposed to postmortems, realized outcomes, and future evidence only for training-set deals, never for test-set deals. Accordingly, hindsight-based supervision cannot inflate test-set performance.

\subsection{Evaluation Metrics}

We predict a distribution over the realized outcome $Y \in \{\SucceedPlus,\FailPlus,\FailMinus\}$. For evaluation, we consider two
binary tasks derived from $Y$:
(i) $\Succeed$/$\Fail$ (deal completes vs.\ terminates), and
(ii) $\Plus$/$\Minus$ (positive vs.\ negative outcome for shareholders).
The $\Plus$/$\Minus$ task can be evaluated against market-implied probabilities (Section~\ref{sec:implied-prob}).

Predicted probabilities over deal outcomes are evaluated using the Brier score \cite{brier1950verification}, a proper scoring rule that incentivizes calibration. The standard Brier score—mean squared error applied to probabilistic forecasts—does not account for class imbalance ($\approx $$84\%$ of deals complete) or the asymmetric risk of merger-arb. 
We therefore report four Brier scores: the standard score $\textbf{Brier}$, a class-balanced variant $\textbf{Brier}_B$
\cite{pratt2024languagemodelsuseforecasting}, a surprise-weighted variant $\textbf{Brier}_S$, and a P\&L-weighted variant $\textbf{Brier}_{\$}$.

\paragraph{Weighted Brier Scores.}
Let $Z_t \in \{0,1\}$ be the binary indicator for the relevant task (e.g., $Z_t=\mathbf{1}\{Y=\SucceedPlus\}$ for $\Succeed$/$\Fail$, or $Z_t~=~\mathbf{1}\{Y~\in~\{\SucceedPlus,\FailPlus\}\}$ for $\Plus$/$\Minus$).
For a forecast $\hat{p}_t \in [0,1]$ of the corresponding positive class, all
weighted variants share the form
$$
\textbf{Brier}_W = \mathbb{E}_{w}\bigl[(Z_t - \hat{p}_t)^2\bigr]
$$
where $\mathbb{E}_w$ denotes a weighted expectation. Weights $w$ are:

\begin{itemize}
    \item \textbf{Class-balanced} ($\textbf{Brier}_B$): inverse class frequency,
    equalizing total weight on each class.
    \item \textbf{Surprise-weighted} ($\textbf{Brier}_S$): $w = 1-p^m_t$ for successes ($Y=\Succeed$), $w = p^m_t$ for failures
    ($Y=\Fail$), emphasizing outcomes the market misread.
    \item \textbf{P\&L-weighted} ($\textbf{Brier}_{\$}$): $w = \frac{S^+_t - S_t}{S_t}$
    for successes and $w = \frac{S_t - S^-_t}{S_t}$ for failures, emphasizing the available return for a correct prediction.
\end{itemize}

\noindent
All weighted variants normalize weights so each class contributes equally (50\%)
and clip outlier weights to prevent dominance by extreme cases. We report scores
averaged over forecast dates within each deal, then across deals.

\paragraph{Murphy Decomposition.} LLM forecasts are often poorly calibrated \cite{schoen2024wisdom, alur2025aia}. We apply the Murphy decomposition \cite{murphy1973new} to separate calibration error from resolution. This allows us to distinguish models that assign systematically biased probabilities from those that simply fail to rank deals by risk.

\paragraph{Market Alignment.} We evaluate how closely forecasts $\hat{p}_t$ track market-implied probabilities $p^m_t$ via mean squared deviation $\boldsymbol{(\hat{p}_t - p^m_t)^2}$ and Pearson correlation $\boldsymbol{\rho(\hat{p}_t, p^m_t)}$. These distinguish models that replicate market assessments from those introducing systematic deviations.
\paragraph{Days to completion.}
The number of days to deal completion is evaluated using \emph{mean absolute percentage error} (\textbf{MAPE}) with respect to the realized number of days, a standard scale-free accuracy measure \citep{flores1986pragmatic}.

\section{Approach}

\begin{figure}
  \includegraphics[width=\columnwidth, trim=0.4in 3.25in 0.4in 0, clip]{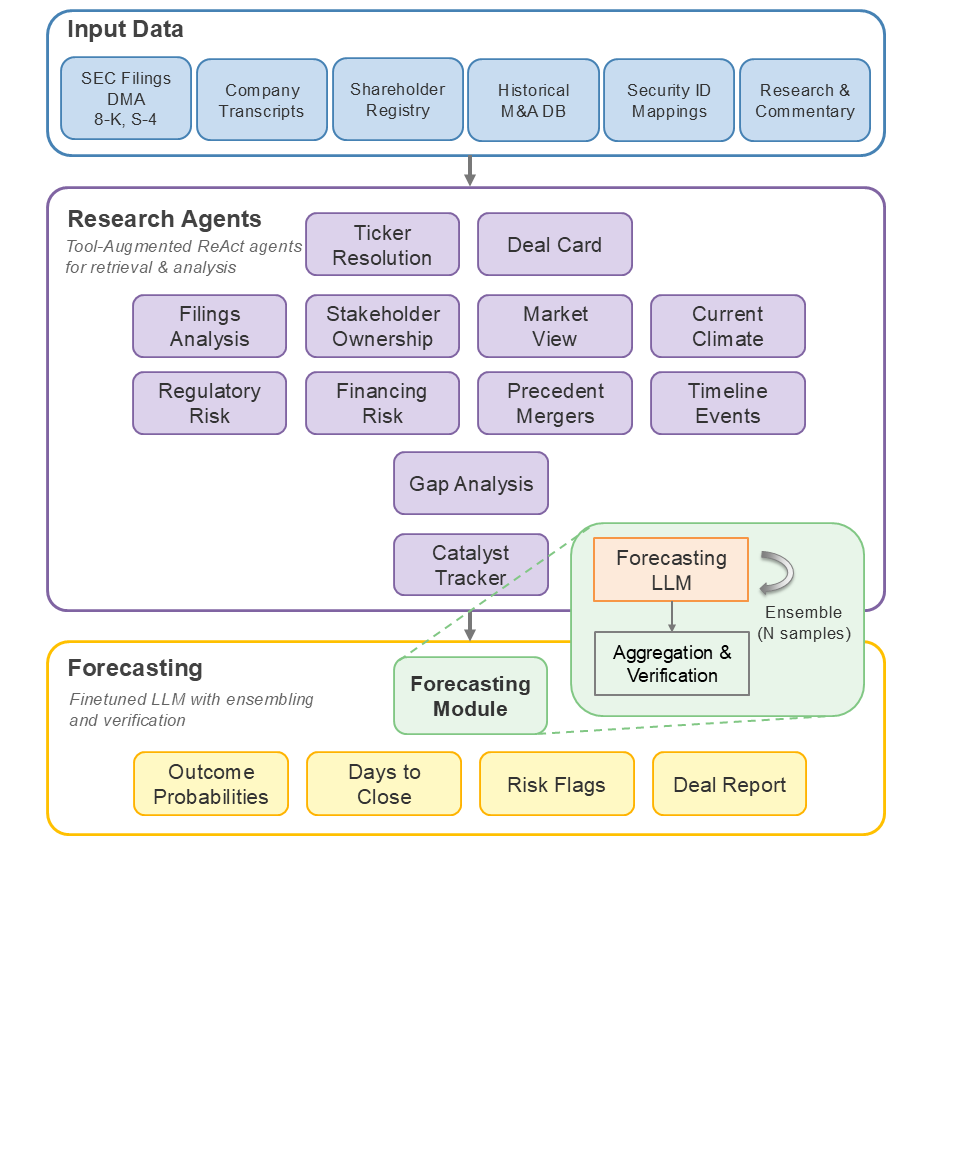}
  \caption{System architecture combining 12 specialized, ReAct research agents with an ensembled, fine‑tuned forecasting LLM.}
  \label{fig:system}
\end{figure}

\subsection{Architecture}

Illustrated in Fig.~\ref{fig:system}, our approach comprises 12 tool-augmented ReAct \cite{yao2022react} agents, each responsible for gathering and analyzing a specific dimension of deal context, together with a forecasting module based on an ensembled, finetuned LLM that synthesizes probabilistic forecasts and a final report.
This two-stage design separates \emph{information retrieval} (research agents) from \emph{probabilistic reasoning} (forecasting module), facilitating independent optimization of each stage while ensuring interpretable intermediate representations.

\begin{table*}[t]
  \caption{Research agents and their functions in execution order (Phases 1--4). Each agent collects and analyzes specific contextual information about the M\&A transaction, which is subsequently integrated by the forecasting agent.}
  \label{tab:research-agents}
  \centering
  \scriptsize
  \begin{tabularx}{\textwidth}{@{}c l X@{}}
    \toprule
    \textbf{Phase} & \textbf{Research Agent} & \textbf{Responsibilities} \\
    \midrule
    1 & Ticker Resolution & Maps company names to unique identifiers across multiple financial data systems (security IDs, entity IDs, organization IDs), enabling cross-platform data retrieval for both acquirer and target companies. \\
    \cmidrule(lr){2-3}
     & Deal Card & Extracts structured deal metadata from SEC filings and expert commentary, including parties, structure (merger type, consideration components), key dates, regulatory requirements, termination fees, and voting thresholds.  \\
    \midrule
    2 & Filings Analysis & Analyzes SEC filings (merger agreements, proxy statements, 8-Ks, S-4s) to extract key facts, identify red flags (adverse terms, litigation risks) and green flags (deal protection provisions), and flag unusual or notable clauses. \\
    \cmidrule(lr){2-3}
     & Stakeholder Ownership & Retrieves and analyzes major shareholder positions ($\geq$5\% and $\geq$10\% thresholds) for both acquirer and target, identifying potential voting dynamics, activist involvement, ownership concentration, and alignment of incentives. \\
    \cmidrule(lr){2-3}
     & Market View & Reviews expert commentary and market sentiment regarding the deal, examining pre-deal health of both parties, industry trends, rumors, and opinions on deal rationale and synergies. \\
    \cmidrule(lr){2-3}
     & Current Climate & Assesses the broader market, regulatory, and economic environment relevant to the transaction, including antitrust climate, interest rate environment, sector-specific headwinds or tailwinds, and geopolitical factors. \\
    \cmidrule(lr){2-3}
     & Regulatory Risk & Analyzes antitrust and regulatory approval risks, international regulatory requirements, industry concentration concerns, and historical precedent for similar deals. \\
    \cmidrule(lr){2-3}
     & Financing Risk & Assesses risks related to deal financing and acquirer's ability to complete the transaction, including debt financing availability, equity raise requirements, credit rating implications, covenant compliance, and acquirer financial health. \\
    \cmidrule(lr){2-3}
     & Precedent Mergers & Identifies and analyzes similar historical M\&A transactions from deal databases, filtering by industry, geography, deal size, and time period. Ranks precedents by relevance and extracts completion rates, timelines, and key risk factors. \\
    \cmidrule(lr){2-3}
     & Timeline Events & Extracts and organizes key events in the deal lifecycle (regulatory filings, shareholder votes, court dates, financing milestones), categorizes by type, and identifies past, ongoing, and future expected catalysts. \\
    \midrule
    3 & Gap Analysis & Performs gap analysis across all previously collected module outputs to identify missing information or overlooked details relevant to deal completion probability, ensuring comprehensive coverage. \\
    \midrule
    4 & Catalyst Tracker & Identifies forward-looking catalysts and near-term developments that could materially impact deal probability or timeline (upcoming regulatory decisions, shareholder vote dates, financing deadlines, competing bid windows). \\
    \bottomrule
  \end{tabularx}
\end{table*}

\subsubsection{Input Data and Research Agents}
\label{sec:input-data}

Input data is obtained from six sources:
\begin{enumerate}
    \item \textbf{Regulatory Filings}: Definitive merger agreements (DMA), proxy statements, S-4 and 8-K filings, etc.
    \item \textbf{Company Transcripts}: Transcripts of company earnings calls and presentations.
    \item \textbf{Shareholder Registry}: Point-in-time institutional stock ownership data.
    \item \textbf{Historical M\&A Database}: Records of past merger attempts, outcomes, and timelines.
    \item \textbf{Security ID Mappings}: Identifiers linking entities and tickers across data systems.
    \item \textbf{Research \& Commentary}: Expert analysis and other research.
\end{enumerate}

Documents are stored in a large-scale vector store built on a domain-finetuned embedding model~\cite{finance_embeds}, together with a finetuned reranker for passage-level relevance. The indexed corpus is intentionally broad rather than restricted to the deals in our evaluation set: it spans global public and private companies, sectors, commodities, macroeconomic developments, and geopolitical events, and comprises millions of documents. The same retrieval stack underlies interactive tools used by human analysts and portfolio managers (and their AI assistants), so agents operate on the same document distribution and interfaces that support real-world forecasting workflows. 

All documents are timestamped. Retrieval is constrained to items released prior to the forecast date to prevent temporal leakage (see Sec.~\ref{sec:leakage}). Agents interact with this corpus through tools that expose full-text search, retrieval, and structured lookups (e.g., holdings, identifiers, and deal attributes). Each research agent is implemented using GPT-4.1 or GPT-5 with access to a subset of these tools and is prompted to extract domain-specific information from retrieved context. 

Agent roles, prompt formulations, and output schemas were iteratively optimized over months of collaboration with veteran merger-arb specialists, who reviewed system outputs on historical deals they had previously covered, compared the automated analysis to their own research, and highlighted gaps for correction. The agents, their responsibilities, and their execution order are summarized in Tab.~\ref{tab:research-agents}.

\subsubsection{Forecasting Module}

The forecasting module uses a designated \emph{forecast LLM} (a finetuned model, or a frontier model such as GPT-5) to convert the research agents' outputs into probabilistic predictions and a final deal report.

\paragraph{Context assembly.}
We first concatenate all agent outputs into a single structured context with citations, and pair this with a forecasting prompt that instructs the forecast LLM to start from historical base rates (completion frequencies), incorporate deal-specific evidence, and explicitly weigh green flags against red flags.
The resulting context is typically from 6.8k-10k tokens (interquartile range), with a median of 8.3k tokens.

\paragraph{Ensemble generation.}
The forecast LLM is then conditioned on this context and sampled $N=5$ times with temperature $0.2$ to produce an ensemble of independent forecasts. Each sample includes outcome probabilities, a forecast of days to completion, and a draft deal report. 

\paragraph{Aggregation \& verification.} For robustness to outliers, we take the median of the $N$ samples as the final prediction for both outcome probabilities and days to completion; \citet{alur2025aia} show that the median of five samples recovers most of the benefit of larger ensembles. To obtain the final deal report, we provide GPT-5 (with tool access) the set of sampled reports. It cross-checks inconsistencies against source documents and produces a single, consolidated, and fact-checked report.

\subsection{Finetuning}
\label{sec:finetuning}

We finetune the \emph{forecast LLM} on historical deals with known outcomes, allowing it to learn patterns that drive merger success or failure and to map qualitative evidence to calibrated forecasts. A central design question is how to construct gold training targets for: (i) the deal report, which is the model's reasoning chain, and (ii) quantitative outputs (probabilities over outcomes and days to completion).

Recent work on reasoning-focused LLMs emphasizes that naive supervision on final outcomes alone is insufficient: models benefit from targets that reflect high-quality \emph{process} as well as correct \emph{answers}, often via teacher distillation and process supervision~\cite{uesato2022process,lightman2023prm,luo2024automatedprm}. Our setting is unusual and arguably more favorable: for each forecast date, we observe the full future evolution of the deal and its ultimate resolution. This allows us to use ex-post information to shape ex-ante gold targets, while still enforcing a strict temporal cutoff on what the model is allowed to see at prediction time. We next detail how we construct these gold targets for both deal reports and outcomes.

\subsubsection{Gold Deal Reports}
\label{sec:gold-reasoning}

\paragraph{Post-mortem analysis.}
For each deal we run a dedicated post-mortem agent with access to full information up to and \emph{after the resolution date} (subject to the overall training cutoff). The post-mortem agent is the only component in our system with access to open-web content, since information leakage is irrelevant once the deal has resolved. It produces three structured artifacts:
\begin{enumerate}
    \item \textbf{General post-mortem}: a narrative of how the deal unfolded and which ex-ante signals (e.g., regulatory posture, shareholder alignment, financing quality) ultimately proved predictive.
    \item \textbf{Timeline post-mortem}: an analysis of the realized sequence of events (filings, regulatory actions, litigation, shareholder votes, competing bids), highlighting which milestones accelerated, delayed, or derailed the deal.
    \item \textbf{Market view post-mortem}: an interpretation of how the market priced the deal over time, including episodes where prices appeared to under- or overreact to new information.
\end{enumerate}

\paragraph{Hindsight-guided deal reports.}
To construct a gold deal report for a specific forecast date $t$, we combine the (pre-$t$) forecasting-module context with the post-mortem artifacts and prompt a teacher model (GPT-5.1) to write a hindsight-guided deal report. We instruct it to: (i) reason \emph{as if} making a forecast at time $t$, (ii) base every claim strictly on information available up to $t$ (with citations), and (iii) use the post-mortem only to identify, in hindsight, which pieces of information \emph{should} have been considered salient at the forecast date, not to introduce future events or facts.

This procedure is inspired by process supervision~\cite{uesato2022process,lightman2023prm,luo2024automatedprm}: the post-mortem plays a similar role to a process reward model, guiding towards reasoning chains that are not only plausible but \emph{causally} informative given the realized path. 
Importantly, even if some future information did leak into the gold deal reports, it would not improve the model's quantitative performance because all evaluations are out-of-sample. 

\subsubsection{Gold Outcome Supervision}
\label{sec:gold-prob}

During training, the realized outcome $Y \in \{\SucceedPlus,\FailPlus,\FailMinus\}$ is known. However, training directly on one-hot labels encourages overconfident, brittle forecasts, especially when the deal is far from resolution. Instead, we construct smoothed three-way gold targets $p_t^{*}(\cdot)$ that blend hard labels with market-implied probability $p^m_t$. 

Recall that $p^m_t$ is defined for the $\Plus$/$\Minus$ task and does not distinguish between the two positive outcomes $\SucceedPlus$ and $\FailPlus$. We therefore build $p_t^{*}(\cdot)$ in two steps.

\paragraph{Negative outcome.}
We first define the gold probability of a negative outcome ($\FailMinus$) as
$$
p_t^{*}(\FailMinus)
= \alpha_t \bigl(1 - p^m_{t+7}\bigr)
+ (1 - \alpha_t)\,\mathbf{1}\{Y=\FailMinus\}
$$
where $\alpha_t \in [0,1]$ is a time-varying smoothing parameter and $p^m_{t+7}$ is the 7-day \emph{forward-shifted} market-implied probability of a $\Plus$ outcome.\footnote{We use a short forward shift so that $p^m_{t+7}$ reflects how the market has digested information around date $t$, rather than the often noisy trading immediately after announcement.}
We set $\alpha_t$ to $0.3$ at announcement and decay it smoothly to zero, so that targets rely more on market priors when uncertainty is high and more on the realized outcome as the deal approaches resolution.

\paragraph{Positive outcomes.}
The remaining probability mass, $1 - p_t^{*}(\FailMinus)$, must be split between the two positive outcomes $\SucceedPlus$ and $\FailPlus$.
The market-implied probability $p^m_t$ only defines their sum, not the allocation. To determine this split, we use the teacher
model (GPT-5) with access to post-mortem artifacts (Sec.~\ref{sec:gold-reasoning}),
which produces a relative distribution over $\SucceedPlus$ and $\FailPlus$.

\paragraph{Days to completion.} We use actual days to completion as the gold target.

\paragraph{Implementation Details}

As the forecast LLM we select GPT-4o, finetuned via the OpenAI API. To help balance the dataset, we oversample terminated deals with weights determined on the validation set.

\section{Results}

\label{sec:results}

\begin{table*}[t]
\caption{Overall results on $\Plus/\Minus$ outcome prediction. 
In all tables, superscripts indicate two-sided paired bootstrap significance of the difference relative to \textsc{Ours (GPT-4o-FT)} ($^\ddagger p < 0.01$, $^\dagger p < 0.05$, $^\circ p < 0.10$).}
\label{tab:evaluation-metrics}
\centering
\small
\begin{tabularx}{\textwidth}{@{} r X lllllllcc @{}}
\toprule
\textbf{\#}  & \textbf{Model}  & $\mathbf{Brier_B}$ $\downarrow$  & $\text{Cal}_B$ $\downarrow$  & $\text{Disc}_B$ $\downarrow$  & $\mathbf{Brier_S}$ $\downarrow$  & $\mathbf{Brier_\$}$ $\downarrow$  & \textbf{MAPE} $\downarrow$  &  $\boldsymbol{(\hat{p}_t - p^m_t)^2}$  & $\boldsymbol{\rho(\hat{p}_t, p^m_t)}$ \\
\midrule
1 & Market $p^m$ & $0.229^{\ddagger}$ & $0.046$ & $0.183$ & $0.454^{\ddagger}$ & $0.319^{\ddagger}$ & $-$ & $-$ & $-$ \\
2 & Market (Platt) & $0.199^{\ddagger}$ & $\mathbf{0.016}$ & $0.183$ & $0.301^{\ddagger}$ & $0.240^{\ddagger}$ & $-$ & ${0.063}$ & ${1.000}$ \\
3 & XGBoost & $\underline{0.186^{\dagger}}$ & $\underline{0.035}$ & $0.153$ & $0.280^{\ddagger}$ & $0.246^{\ddagger}$ & $-$ & $0.064$ & $0.753$ \\
4 & Claude-37-Sonnet & $0.259^{\ddagger}$ & $0.146$ & $0.108$ & $0.276^{\ddagger}$ & $0.267^{\ddagger}$ & $0.433^{\circ}$ & $0.156$ & $0.291$ \\
5 & Gemini-3-Flash & $0.250^{\ddagger}$ & $0.144$ & $\mathbf{0.101}$ & $0.289^{\ddagger}$ & $0.282^{\ddagger}$ & $0.423^{\circ}$ & $0.158$ & $0.319$ \\
6 & Gemini-3-Pro & $0.231^{\ddagger}$ & $0.118$ & ${0.110}$ & $0.276^{\ddagger}$ & $0.268^{\ddagger}$ & $0.497^{\dagger}$ & $0.149$ & $0.339$ \\
7 & GPT-5.1 & $0.214^{\ddagger}$ & $0.110$ & $\underline{0.103}$ & $0.244^{\ddagger}$ & $0.245^{\ddagger}$ & $0.465^{\ddagger}$ & $0.140$ & $0.355$ \\
8 & GPT-4o & $0.201^{\ddagger}$ & $0.089$ & $0.111$ & ${0.226^{\dagger}}$ & ${0.234^{\ddagger}}$ & $0.463^{\ddagger}$ & $0.133$ & $0.299$ \\
9 & GPT-4o + Isotonic & $\underline{0.186^{\ddagger}}$ & $0.075$ & $0.110$ & $\underline{0.219^{\ddagger}}$ & $\underline{0.228^{\ddagger}}$ & $0.463^{\dagger}$ & $0.149$ & $0.314$ \\
10 & Ours (GPT-4o-FT) & $\mathbf{0.151}$ & $0.039$ & $0.112$ & $\mathbf{0.187}$ & $\mathbf{0.178}$ & $\mathbf{0.385}$ & $0.139$ & $0.358$ \\

\bottomrule

\end{tabularx}
\end{table*}

\begin{table}[t]
\caption{Overall results on $\Succeed$/$\Fail$ prediction.}
\label{tab:evaluation-metrics-completion}
\centering
\small
\begin{tabularx}{\columnwidth}{@{} r X ccc @{}}
\toprule
\textbf{\#}  & \textbf{Model}  & $\mathbf{Brier_B}$ $\downarrow$  & $\text{Cal}_B$ $\downarrow$  & $\text{Disc}_B$ $\downarrow$ \\
\midrule
1 & XGBoost             & $\underline{0.164^{\ddagger}}$ & $\mathbf{0.013}$ & $0.151$         \\
2 & Claude-37-Sonnet    & $0.239^{\ddagger}$             & $0.135$       & $\mathbf{0.099}$   \\
3 & Gemini-3-Flash      & $0.227^{\ddagger}$             & $0.118$       & $0.107$            \\
4 & Gemini-3-Pro        & $0.210^{\ddagger}$             & $0.101$       & $0.109$            \\
5 & GPT-5.1             & $0.173^{\ddagger}$             & $0.065$       & $\underline{0.106}$ \\
6 & GPT-4o              & $0.177^{\ddagger}$             & $0.057$       & $0.118$            \\
7 & Ours (GPT-4o-FT)    & $\mathbf{0.126}$               & $0.019$       & $0.107$            \\
\bottomrule
\end{tabularx}
\end{table}

\begin{table}[t]
\caption{Qualitative example. Each scenario modifies the baseline context; $\Delta \hat{p}_t$ is averaged over 5 predictions.}
\label{tab:scenarios}
\centering
\small
\begin{tabularx}{\columnwidth}{X cc r}
\toprule
Scenario & Expected & $\hat{p}_t$ & $\Delta\hat{p}_t$ \\
\midrule
Baseline (no event) & -- & 82\% & -- \\
ISS recommends FOR & $\uparrow$ & 88\% & +5\% \\
ISS recommends AGAINST & $\downarrow$ & 73\% & $-$9\% \\
Major shareholder support & $\uparrow$ & 87\% & +4\% \\
Major shareholder opposition & $\downarrow$ & 54\% & $-$28\% \\
Bidder raises offer 10\% & $\uparrow$ & 91\% & +9\% \\
Rumors of financing concerns & $\downarrow$ & 81\% & $-$1\% \\
\bottomrule
\end{tabularx}

\end{table}

\label{sec:baselines}

\paragraph{Baselines.}
We compare our finetuned system to a variety of frontier models with access to identical context, plus baselines using market and deal-level features:
\begin{itemize}
    \item \textbf{Market-implied probability.} $p^m_t$ from Sec.~\ref{sec:implied-prob}.
    \item \textbf{Calibrated market-implied probability.} Platt-scaled $p^m_t$ using train+validation data.
    \item \textbf{XGBoost on structured features.} Using $p^m_t$ and 26 deal-specific features, including arbitrage profit, deal value, ownership structure, consideration mix, and competitive dynamics (Appendix ~\ref{app:xgboost-details}).
\end{itemize}

\paragraph{Overall $\Plus$/$\Minus$ outcome prediction.} 
As reported in Tab. \ref{tab:evaluation-metrics}, our finetuned system (GPT-4o-FT) achieves the lowest Brier scores across all three weighting schemes, outperforming second-place models XGBoost on class-balanced Brier score ($\mathbf{Brier}_B$, 0.151 vs 0.186), and second-place model GPT-4o on surprise and P\&L weighted Brier scores ($\mathbf{Brier_S}$, 0.187 vs. 0.226; $\mathbf{Brier_\$}$, 0.178 vs. 0.234). The surprise and P\&L weighting schemes expose XGBoost's reliance on the market-implied probability feature--when the market is surprised, the XGBoost model performs poorly. In contrast to XGBoost, our system has low correlation with market implied probability (0.358 vs. 0.753).

\paragraph{Finetuning substantially improves performance.}
Even with access to curated context from our research agents, frontier models including Gemini-3-Pro, GPT-5.1, and GPT-4o underperform XGBoost on $\mathbf{Brier}_B$ (Tab.~\ref{tab:evaluation-metrics}, rows 6--8 vs.~3). Murphy decomposition suggests that these models achieve stronger \textit{discrimination} (\textbf{Disc}) than XGBoost---they better separate higher- and lower-risk deals---but have weaker \textit{calibration} (\textbf{Cal}). Finetuning substantially improves performance, especially calibration (row 10 vs.~8), resulting in the best overall Brier score. We observe similar patterns for $\Succeed$/$\Fail$ prediction (Tab.~\ref{tab:evaluation-metrics-completion}), where our system again achieves the lowest Brier scores by a large margin. 

\paragraph{Finetuning outperforms post-hoc calibration.}
The improvement from finetuning does not arise from post-hoc calibration alone: isotonic regression \cite{niculescu2005predicting} improves the class-balanced Brier score of GPT-4o from 0.201 to 0.186 (row 9 vs.~8), but still falls well short of GPT-4o-FT at 0.151. Platt and temperature scaling similarly fail to close the gap (see Appendix~\ref{app:calibration}). 
While Murphy discrimination is broadly unchanged after finetuning, we note that it is a relatively coarse diagnostic (computed with 10 forecast bins), and may be missing finer-grained improvements in predictive quality. 

\paragraph{Finetuning improves days-to-close prediction.}
Our system achieves a median absolute percentage error (MAPE) of 0.385 on completed deals, compared to 0.463 for GPT-4o.

\paragraph{Qualitative example.}
To illustrate how the model adjusts probabilities in the response to information, in Tab. \ref{tab:scenarios} we intervene on an example deal, modifying the deal context by hand.
Across six example scenarios spanning three causal mechanisms--proxy advisory influence (ISS recommendation), shareholder voting, and deal terms--in each case the model adjusts $\hat{p}_t$ in the expected direction. The largest effect (shareholder opposition) reflects a realistic constraint.

\subsection{Ablation studies}
\label{sec:ablations}

In Tab. \ref{tab:ablations} we report ablations to evaluate how much each component supports the system. 

\paragraph{Scaling the training set is likely to yield further gains.}
Training on only 50\% of the data (row 2) -- stratified by region to maintain diversity -- degrades $\textbf{Brier}_B$ from 0.151 to 0.177, suggesting that performance has not yet saturated with respect to training set size.

\paragraph{Dataset balancing requires careful tuning.}
During training we oversampling terminated deals, multiplying the frequency of terminated negatives and positives by 1.65 and 1.25,
respectively. Removing oversampling during finetuning (row 3) degrades $\textbf{Brier}_B$ from 0.151 to 0.170. Oversampling too aggressively, with weights (2.0, 1.65) degrades $\textbf{Brier}_B$ to 0.163 (row 4).

\paragraph{Hindsight guidance and market smoothing provide gains.}
We also isolate the contribution of our supervision recipe. Removing hindsight-guided postmortem supervision (row 5) degrades $\mathbf{Brier}_B$ from 0.151 to 0.183, while removing market-probability smoothing (row 6) degrades it to 0.181.

\paragraph{Specialized research agents add value.}
Ablating the information provided by specialized research agents (Deal Card only, row 7) degrades $\textbf{Brier}_B$ from 0.151 to 0.172. In rows 8--14 we add back one research agent at a time (corresponding to Tab. \ref{tab:research-agents}), in each case observing improvements over Deal Card only (row 7) that indicate that each agent contributes useful information. Since research agent responsibilities are informed by merger-arb specialists, we do not enforce that agents retrieve non-overlapping information; nevertheless we find that removing agents individually (Appendix~\ref{app:loo-ablations}) generally worsens performance, although with limited statistical significance.

\subsection{Deal reports}
\label{sec:qualitative-analysis}

We deploy this system primarily as a decision-support tool for discretionary investors, making the reasoning and grounding of the reports especially important. In this section, we study the quality of the generated deal reports.

\paragraph{Grounding.}
Using an LLM-based grader, we assess the factual grounding of generated deal reports by: (1) identifying claims in the report, and (2) assessing the level of evidence for each claim. Our finetuned (GPT-4o-FT) model averages 33 claims per report with an unsupported-claim rate of 0.1\%, compared with 9 claims and 0.3\% for the GPT-4o base model, suggesting that finetuning with hindsight-guided supervision does not introduce hallucinations or reduce factual grounding.

\paragraph{Rubric-based assessment.}
A key application of our system is forming a Day-1 view in response to a deal announcement. To evaluate whether our reports capture the resulting decision-relevant issues within hours of announcement, we construct deal-specific checklists of the key ex-ante risks and mitigants visible at the initial forecast date and score whether each report identifies them (rubric construction and grading details in Appendices~\ref{app:rubric-curation} and~\ref{app:rubric-grading}). GPT-4o-FT achieves $50.2\%$ rubric coverage vs.\ $21.5\%$ for the base model (${+}28.7$ pp,  $p{<}0.001^{***}$). Coverage remains partial even for the finetuned model, reflecting the difficulty of identifying non-obvious, analyst-level, and idiosyncratic insights from Day-1 context.

\paragraph{Failure analysis.}
Appendix~\ref{app:error-analysis} analyzes the largest GPT-4o-FT forecasting errors. Most arise from missing forecast-date context rather than faulty reasoning. Web search would reduce this gap, but we disable it to avoid information leakage (Section~\ref{sec:leakage}).

\begin{table}[t]
\caption{Impact of system ablations on $\Plus/\Minus$ outcome prediction.}
\label{tab:ablations}
\centering
\small
\begin{tabularx}{\columnwidth}{@{} r X ccc @{}}
\toprule
\textbf{\#}  & \textbf{Model}  & $\mathbf{Brier_B}$ $\downarrow$  & $\text{Cal}_B$ $\downarrow$  & $\text{Disc}_B$ $\downarrow$ \\
\midrule
1  & Ours (GPT-4o-FT)               & $\mathbf{0.151}$               & $0.039$ & $\mathbf{0.112}$ \\
2 & \dots 50\% data                & $0.177^{\ddagger}$             & $0.062$ & $0.115$ \\
3 & \dots No oversampling          & $0.170^{\ddagger}$             & $0.058$ & $\mathbf{0.112}$ \\
4 & \dots More oversampling        & $0.163$                        & $0.042$ & $0.121$ \\
5 & \dots No hindsight guide & $0.183^{\ddagger}$ & $0.062$ & $0.121$ \\
6 & \dots No $p^m_t$ smoothing & $0.181^{\ddagger}$ & $0.081$ & $0.100$ \\
7 & \dots Deal Card only      & $0.172^{\dagger}$              & $\mathbf{0.036}$ & $0.137$ \\
8 & \dots Filings Analysis    & $0.163$                        & $0.040$ & $0.123$ \\
9 & \dots Stakeholder Own.        & $0.158^{\circ}$                & $0.042$ & $0.117$ \\
10 & \dots Market View         & $0.165^{\dagger}$              & $0.043$ & $0.122$ \\
11 & \dots Current Climate     & $0.170^{\circ}$                & $0.040$ & $0.130$ \\
12 & \dots Precedent Mergers     & $0.175^{\dagger}$              & $\mathbf{0.036}$ & $0.140$ \\
13 & \dots Regulatory Risk    & $0.167$                        & $0.039$ & $0.128$ \\

\bottomrule
\end{tabularx}
\end{table}

\section{Conclusion}

We present an LLM-based merger-arbitrage forecasting system that combines specialist research agents with hindsight-guided finetuning. It outperforms frontier models, market-implied probabilities, and XGBoost on held-out deals; ablations trace the gains to specialist context, hindsight-guided supervision, target smoothing, class balancing, and training-set scale.

\newpage

\section*{Impact Statement}
This paper studies language-model-based forecasting for announced public M\&A transactions. The main positive impact is improved decision support: the system may help analysts process complex public information more efficiently and consistently. Potential risks include over-reliance on model outputs, increased informational asymmetry, and misuse in trading contexts. Our system is intended for human decision support, not autonomous trading, and should be used with appropriate oversight.

\bibliography{custom}
\bibliographystyle{icml2026}
\newpage
\appendix
\onecolumn

\section{Hyperparameters}
\label{implementation-details}

\subsection{Forecasting LLM}
\label{fine-tuning-details}

We fine-tuned a supervised forecasting model using the OpenAI supervised fine-tuning API on \texttt{gpt-4o-2024-08-06}. 

Table~\ref{tab:ft-details} contains fine-tuning and inference settings. 

\begin{table}[h]
\centering
\small
\begin{tabularx}{\linewidth}{@{} >{\raggedright\arraybackslash}p{0.33\linewidth} X @{}}
\toprule
\textbf{Item} & \textbf{Value} \\
\midrule
Fine-tuning method & OpenAI supervised fine-tuning API \\
Base model & \texttt{gpt-4o-2024-08-06} \\
Teacher model for gold outputs & \texttt{gpt-5.1} \\
System prompt & ``You are a superforecaster with merger arbitrage expertise.'' \\
Epochs & 1 \\
Batch size & 2 \\
Learning rate multiplier & 1 \\
Inference temperature & 0.2 \\
Inference top-$p$ & 1.0 \\
Inference max tokens & 16384 \\
\bottomrule
\end{tabularx}
\caption{Fine-tuning and inference settings for the GPT-4o supervised finetuned model.}
\label{tab:ft-details}
\end{table}

\subsection{XGBoost Baseline}
\label{app:xgboost-details}

Our strongest structured baseline is a three-class XGBoost classifier trained on point-in-time structured deal features. 

To reduce class imbalance, inverse-frequency class weights are applied at training time and capped at a maximum ratio of 10$\times$ between the rarest and most common class weights.

The feature set combines market-implied information with point-in-time deal-structure fields, as listed in Table~\ref{tab:xgb-bbg-features}. The XGBoost hyperparameters are included in Table~\ref{tab:xgb-bbg-hparams}.

\begin{table}[h]
\centering
\small
\begin{tabularx}{\linewidth}{@{} >{\raggedright\arraybackslash}p{0.28\linewidth} X @{}}
\toprule
\textbf{Feature group} & \textbf{Variables} \\
\midrule
Market-implied features &
market-implied probability of success, normalized merger spread, normalized downside-to-break risk \\
Economics / valuation &
announcement premium, current premium, log transaction value, arbitrage profit estimate \\
Ownership / control &
fraction of target shares sought, target shareholder ownership in the post-merger entity \\
Consideration mix &
cash component present, stock component present, contingent consideration present, cash share of total consideration \\
Legal / process terms &
financing condition present, MAE clause present, dissenters' rights present, go-shop period length, drop-dead date present \\
Bid dynamics &
friendly bid indicator, hostile bid indicator, unsolicited bid indicator \\
Deal type indicators &
acquisition indicator, merger indicator, divestiture indicator \\
Termination fees &
acquirer-to-target termination fee / deal value, target-to-acquirer termination fee / deal value \\
\bottomrule
\end{tabularx}
\caption{Features used by the \texttt{XGBoost} baseline.}
\label{tab:xgb-bbg-features}
\end{table}

\begin{table}[h!]
\centering
\small
\begin{tabular}{@{} ll @{}}
\toprule
\textbf{Hyperparameter} & \textbf{Value} \\
\midrule
Objective & \texttt{multi:softprob} \\
Number of classes & 3 \\
Max depth & 5 \\
Min child weight & 3 \\
Gamma & 0.1 \\
Learning rate & 0.05 \\
Number of trees & 50 \\
Subsample & 0.8 \\
Column subsample & 0.8 \\
Optimization metric & \texttt{mlogloss} \\
Missing-value handling & Training-median imputation for numeric features \\
Class weighting & Inverse-frequency weights \\
\bottomrule
\end{tabular}
\caption{Hyperparameters for XGBoost, tuned on the validation set.}
\label{tab:xgb-bbg-hparams}
\end{table}

\section{Additional Results}

\subsection{Additional post-hoc calibration baselines}
\label{app:calibration}

We include additional calibration baselines in Table \ref{tab:extra-calib}.

\begin{table*}[h]
\caption{Additional post-hoc calibration baselines for GPT-4o on $\Plus/\Minus$ outcome prediction.}
\centering
\small
\begin{tabularx}{\textwidth}{@{} X lllllllcc @{}}
\toprule
\textbf{Model}  & $\mathbf{Brier_B}$ $\downarrow$  & $\text{Cal}_B$ $\downarrow$  & $\text{Disc}_B$ $\downarrow$  & $\mathbf{Brier_S}$ $\downarrow$  & $\mathbf{Brier_\$}$ $\downarrow$  & \textbf{MAPE} $\downarrow$  &  $\boldsymbol{(\hat{p}_t - p^m_t)^2}$  & $\boldsymbol{\rho(\hat{p}_t, p^m_t)}$ \\
\midrule
GPT-4o + Platt & $0.214^{\ddagger}$ & $0.101$ & $0.111$ & $0.247^{\ddagger}$ & $0.256^{\ddagger}$ & $0.463^{\dagger}$ & $0.148$ & $0.302$ \\
GPT-4o + Temp-Scale & $0.222^{\ddagger}$ & $0.109$ & $0.111$ & $0.254^{\ddagger}$ & $0.263^{\ddagger}$ & $0.463^{\dagger}$ & $0.149$ & $0.300$ \\
\bottomrule
\end{tabularx}
\label{tab:extra-calib}
\end{table*}

\subsection{Selected leave-one-out ablations}
\label{app:loo-ablations}

We include leave-one-out ablations in Table \ref{tab:loo-ablations}.

\begin{table*}[h]
\caption{Selected leave-one-out ablations from the full GPT-4o-FT system on $\Plus/\Minus$ outcome prediction.}
\centering
\small
\begin{tabularx}{\textwidth}{@{} X lllllllcc @{}}
\toprule
\textbf{Model}  & $\mathbf{Brier_B}$ $\downarrow$  & $\text{Cal}_B$ $\downarrow$  & $\text{Disc}_B$ $\downarrow$  & $\mathbf{Brier_S}$ $\downarrow$  & $\mathbf{Brier_\$}$ $\downarrow$  & \textbf{MAPE} $\downarrow$  &  $\boldsymbol{(\hat{p}_t - p^m_t)^2}$  & $\boldsymbol{\rho(\hat{p}_t, p^m_t)}$ \\
\midrule
FT w/o Financing Risk & $0.155$ & $0.037$ & $0.118$ & $0.206$ & $0.194$ & $0.405$ & $0.137$ & $0.373$ \\
FT w/o Filings Analysis & $0.156$ & $0.036$ & $0.120$ & $0.190$ & $0.193$ & $0.404$ & $0.136$ & $0.366$ \\
FT w/o Stakeholder Own. & $0.158$ & $0.043$ & $0.114$ & $0.201$ & $0.188$ & $0.393$ & $0.138$ & $0.352$ \\
FT w/o Deal Card              & $0.165$ & $0.048$ & $0.117$ & $0.202$ & $0.195$ & $0.397$ & $0.140$ & $0.354$ \\
  FT w/o Current Climate        & $0.156$ & $0.038$ & $0.120$ & $0.201$ & $0.188$ & $0.412$ & $0.136$ &
  $0.369$ \\                                                                                                 
  FT w/o Precedent Mergers      & $0.161$ & $0.043$ & $0.117$ & $0.208$ & $0.195$ & $0.400$ & $0.138$ &
  $0.361$ \\                                                                                                 
  FT w/o Market View            & $0.154$ & $0.039$ & $0.115$ & $0.204$ & $0.194$ & $0.396$ & $0.138$ &
  $0.361$ \\                                       FT w/o Gap Analysis            & $0.164$ & $0.043$ & $0.120$ & $0.213$ & $0.202$ & $0.413$ & $0.141$ & $0.336$ \\                           FT w/o Regulatory Risk        & $0.157$ & $0.036$ & $0.120$ & $0.207$ & $0.190$ & $0.415$ & $0.136$ &
  $0.371$ \\                                                                                                              
\bottomrule
\end{tabularx}
\label{tab:loo-ablations}
\end{table*}

\clearpage
\section{Qualitative Evaluation}
\label{app:qualitative-eval}

This appendix details the rubric construction and grading procedure used in the rubric-based assessment of Section~\ref{sec:qualitative-analysis}, followed by an error analysis of the worst forecasting errors made by GPT-4o-FT on the test set.

\subsection{Day-1 Rubric Curation}
\label{app:rubric-curation}

For each deal in the test set, we generate a Day-1 rubric by prompting \texttt{gpt-5} with the deal-announcement context and access to a set of research tools. The model is asked to enumerate  non-obvious risks, mitigants, or deal-specific facts that were discoverable on or shortly after the announcement date; research published within roughly one week of announcement is admissible, but facts that only became public later are excluded unless there were documented signals at announcement.

\subsection{Rubric Grading}
\label{app:rubric-grading}

Each model report is graded against its rubric by \texttt{gpt-5} The grader assigns each rubric item one of three labels (Fully / Partially / Not Covered) and returns a coverage score in $[0,1]$ together with a short explanation. The score reported as ``rubric coverage'' in Section~\ref{sec:qualitative-analysis} is the mean of this per-deal coverage score over the test set.

\subsection{Error Analysis}
\label{app:error-analysis}

We take the 100 worst forecasting errors on the test set (largest squared-error samples) and classify each with \texttt{gpt-5} into one of five primary root causes: (1) \textsc{Missing, stale, or incorrect context} (the prompt was missing or stale on a \textbf{decision-relevant} fact), (2) \textsc{Missed Signal} {key fact(s) were in the prompt but the model under-weighted it), (3) \textsc{Miscalibration} (the model identified the right risks but assigned the wrong probability), (4) \textsc{Hallucination} (the model introduced a fact not in the prompt and not true), and (5) \textsc{Structural Limit} (the deal outcome was driven by an exogenous shock -- e.g., a geopolitical shock or disease outbreak -- that could not plausibly have anticipated.). Table~\ref{tab:error-distribution} reports the resulting distribution.

\begin{table}[h]
\centering
\small
\begin{tabular}{lrr}
\toprule
\textbf{Primary root cause} & \textbf{N} & \textbf{\%} \\
\midrule
\textsc{Missing, stale, or incorrect context}   & 60 & 60.0 \\
\textsc{Missed signal provided in the prompt}     & 20 & 20.0 \\
\textsc{Miscalibration}     & 17 & 17.0 \\
\textsc{structural limit}  &  3 &  3.0 \\
\textsc{Hallucination}  &  0 &  0.0 \\

\midrule
\bottomrule
\end{tabular}
\caption{Primary root cause of the 100 worst forecasting errors made by GPT-4o-FT, classified by \texttt{gpt-5}.}
\label{tab:error-distribution}
\end{table}

\paragraph{Missing or stale context examples.}
To illustrate the texture of these gaps, Table~\ref{tab:error-cases} lists ten representative cases from the information-gap subset (N=60). For each case we report the model's predicted probability of a positive outcome, the realized outcome, and a description of the publicly available context that was missing or stale in the prompt at the sample date.

\begin{table}[h]
\centering
\small
\begin{tabularx}{\linewidth}{@{}rlX@{}}
\toprule
\textbf{Pred} & \textbf{Truth} & \textbf{Context missing or stale in the prompt} \\
\midrule
0.98 & Failed    & Missing the tender-offer terms (two-thirds minimum acceptance, offer price). \\
0.95 & Failed    & Missing the appraisal-payout cap that auto-terminates the merger if breached. \\
0.86 & Failed    & Missing that the tender price was set far below the prevailing market price. \\
0.81 & Failed    & Missing a blocking shareholder's $>$30\% stake, which made acceptance implausible. \\
0.75 & Failed    & Missing the target board's public rejection of the proposal before the sample date. \\
0.65 & Failed    & Missing ISS and Glass Lewis recommendations against the deal. \\
\midrule
0.45 & Completed & Missing that both shareholder bodies had already overwhelmingly approved the merger pre-sample. \\
0.20 & Completed & Stale 2019 lapsed bid in the prompt; the live 2025 recommended offer and its approvals were absent. \\
0.19 & Completed & Missing that the offer had been declared unconditional once all regulatory conditions cleared. \\
0.12 & Completed & Missing anchor-shareholder alignment on both sides and confirmation of key early regulatory approvals. \\
0.07 & Completed & Missing the formal merger announcement on the sample date; the prompt framed the deal as an unsigned rumor. \\
\bottomrule
\end{tabularx}
\caption{Ten representative information-gap cases from the worst-100 set. In each case the model's reasoning is internally coherent given the prompt, but the prompt itself is missing a single decision-relevant fact that would have flipped the conclusion.}
\label{tab:error-cases}
\end{table}

\end{document}